\documentclass[11pt]{article}

\usepackage[preprint]{acl}

\usepackage{times}
\usepackage{latexsym}

\usepackage[T1]{fontenc}
\usepackage[utf8]{inputenc}

\usepackage{microtype}

\usepackage{inconsolata}
\usepackage{amsmath}
\usepackage{amssymb}
\usepackage{algorithm}
\usepackage{algpseudocode}
\usepackage{fvextra}
\usepackage{booktabs}
\usepackage{tabularx}
\usepackage{array}
\usepackage{ragged2e}
\usepackage{tikz}
\usepackage{xurl}
\usepackage{xspace}
\usepackage{mathtools}
\usepackage{graphicx}
\usepackage{subcaption}
\usetikzlibrary{arrows.meta, positioning, shapes.geometric, fit}
\usepackage{multirow}
\usepackage{bbm}

\usepackage{graphicx}

\DefineVerbatimEnvironment{CodeBlock}{Verbatim}{breaklines=true,breakanywhere=true,fontsize=\small}
\RecustomVerbatimEnvironment{verbatim}{Verbatim}{breaklines=true,breakanywhere=true,fontsize=\small}

\newcommand{\tabresize}[1]{\resizebox{0.88\textwidth}{!}{#1}}
\newcommand{\AlgInput}{\Statex \textbf{Input:} }
\newcommand{\AlgOutput}{\Statex \textbf{Output:} }
\newcommand{\Exe}{\textsc{Exe}\xspace}
\newcommand{\Evo}{\textsc{Evo}\xspace}
\newcommand{\Rfc}{\textsc{Rfc}\xspace}
\newcommand{\Rfn}{\textsc{Rfn}\xspace}
\newcommand{\Ext}{\textsc{Ext}\xspace}
\newcommand{\MetaEvo}{\textsc{MetaEvo}\xspace}
\newcommand{\Gate}{\textsc{Gate}\xspace}
\newcommand{\Sample}{\textsc{Sample}\xspace}
\newcommand{\Filter}{\textsc{Filter}\xspace}
\newcommand{\release}{\textsc{release}\xspace}
\title{You Live More Than Once: Towards Hierarchical Skill Meta-Evolving}

\author{
  \textbf{Xujun Li\textsuperscript{1}\textsuperscript{*}}\ \ 
  \textbf{Kehan Zheng\textsuperscript{1}\textsuperscript{*}}\ \ 
  \textbf{Mingyuan Zhao\textsuperscript{1}}\ \ 
  \textbf{Yize Geng\textsuperscript{1}}\ \ 
  \textbf{Jinfeng Zhou\textsuperscript{1}}
  \\
  \textbf{Qi Zhu\textsuperscript{2}}\ \ 
  \textbf{Fei Mi\textsuperscript{2}}\ \ 
  \textbf{Lifeng Shang\textsuperscript{2}}\ \ 
  \textbf{Minlie Huang\textsuperscript{1}}\ \ 
  \textbf{Hongning Wang\textsuperscript{1}{\textsuperscript\textdagger}}
  \\
  \textsuperscript{1}Department of Computer Science and Technology, Tsinghua University
  \\
  \textsuperscript{2}Huawei Foundation Model Department
  \\
  \texttt{lixujun20@gmail.com}\ \ \ \ \ \ \texttt{hw-ai@tsinghua.edu.cn}
}

\begin{document}
\maketitle

{\let\thefootnote\relax
\footnotetext{\textsuperscript{*}\ Equal contribution.}}

{\let\thefootnote\relax
\footnotetext{\textdagger\ Corresponding author.}}

\begin{abstract}
Test-time skill evolving is regarded as a new paradigm for enhancing deployed agentic systems. Existing works mainly focus on hard-coded skill evolving strategies or parametric learning that rely on expensive parameter updates in the underlying LLMs. 
In this paper, we demonstrate that test-time refinement of the skill evolving framework itself is necessary for continuous improvement of the agent systems in different downstream scenarios, and lightweight algorithmic adaptation is feasible. 
Specifically, we propose HiSME, a lightweight hierarchical skill meta-evolving solution that jointly optimizes skills and the skill evolving strategy by learning meta-skills from agents' task execution traces. 
Experiments on mainstream agentic benchmarks show that meta-evolving can produce a higher-quality skill library than pure skill evolving and can derive diverse meta-skills for different scenarios, thereby facilitating future continual experience learning. Our code is temporarily public at \url{https://anonymous.4open.science/r/HiSME-BD45}.

\end{abstract}

\section{Introduction}

Skills have recently been introduced as plug-and-play modules for large language model-based agents. 
As a desired way to extend the manually curated skills, \textbf{skill evolving} refers to a test-time learning paradigm in which, after deployment, an agent improves its behavior by distilling execution traces into reusable skills and maintaining them over time~\citep{trace2skill,skillx,skillforge}. In this sense, skill evolving plays a role analogous to model parameter updates, while remaining lightweight, modular, and explainable in text space.

% Skills have recently emerged as plug-and-play modules that equip large language model (LLM) agents with reusable, modular capabilities. 
% To move beyond manually curated skill sets, \textbf{skill evolving} has been proposed as a test-time learning paradigm in which a deployed agent improves its behavior by distilling execution traces into reusable skills and maintaining them over time~\citep{trace2skill,skillx,skillforge}. 
% Skill evolving thus plays a role analogous to model parameter updates, but stays lightweight, modular, and interpretable, with all changes carried out in text space.

Recent work has begun to develop skill evolving strategies, largely focusing on maintaining a skill library that improves agent performance~\citep{skillx,skillclaw,skilllens,skillmoo,psn,skillsvote}. 
However, in most cases, the evolving strategy itself remains ``static'': it does not accumulate evidence to improve its own extraction, maintenance, or refinement process over the course of deployment. This creates two practical limitations. First, when the skill generation algorithm has a systematic flaw, the agent system must repeatedly spend maintenance effort repairing the same defect pattern. Second, a fixed evolving strategy is unlikely to fit all target domains, where requirements on skill specification, scope, and maintenance may vary substantially or even conflict. For example, data-processing tasks may benefit from general callable functions, whereas safety-critical settings may favor narrow checklist-type skills. Some works construct skill evolving frameworks through parametric learning~\citep{skillrl,skill1,d2skill,xskill,skillmas,openclaw_rl}, which however impose substantial computational overhead in practice.

Therefore, the core question we aim to study is:

\vspace{1mm}

\textit{Can the skill evolving framework itself also be optimized automatically at test time?}

\vspace{1mm}

% Existing skill evolving methods mainly focus on building a skill library that boosts agent performance in downstream tasks \citep{skillx,skillclaw,skilllens,skillmoo,psn,skillsvote}. 
% However, the evolving strategy itself often remains \textit{static}: it does not accumulate evidence to refine its own extraction, maintenance, or refinement procedures during deployment. 
% This static nature creates two practical limitations. 
% First, when the skill execution procedure contains a systematic flaw, the agent must repeatedly repair the same recurring defects. 
% Second, a one-size-fits-all evolving strategy rarely transfers across domains, where requirements on skill granularity, scope, and maintenance can vary substantially or even conflict. For instance, data-processing tasks often benefit from general callable functions, whereas safety-critical settings favor narrow, checklist-style skills. A parallel line of work addresses these issues through parametric meta-learning~\citep{skillrl,skill1,d2skill,xskill,skillmas,openclaw_rl}, but at the cost of substantial computational overhead in practice. This motivates our central research question:
% \vspace{1mm}
% \textit{Can the skill evolving framework itself be optimized automatically at test time?}
% \vspace{1mm}

We answer this question with \textbf{Hierarchical Skill Meta-Evolving} (HiSME), a framework that optimizes both task skills and the skill maintenance process itself by producing and maintaining meta-level experiences. HiSME aims not only to build a reusable skill library, but also to adapt the skill evolving framework itself so that it can keep improving skills for future tasks and domains. In HiSME, all updates are performed at test time in the text space, without modifying the parameters of the underlying LLMs, therefore light-weighted.

% We address this question with \textbf{Hierarchical Skill Meta-Evolving} (HiSME), a framework that jointly optimizes task-level skills and the skill maintenance process itself by producing and accumulating meta-level experiences. 
% Beyond building a reusable skill library, HiSME continually adapts the evolving framework itself, enabling sustained improvement across future tasks and domains. 
% All updates take place at test time in text space, without modifying the parameters of the underlying LLMs, keeping the approach lightweight.

We evaluate HiSME on several agentic benchmarks. 
The results show that HiSME produces stronger skill libraries than static evolving baselines, improves end-to-end agent performance, and learns evolving strategies that surpass the framework's initial capability. 
These findings suggest that skill evolving can benefit from optimizing not only the skills being created, but also the process that creates and maintains them.

Our contributions are as follows:

\begin{itemize}
    \item We identify the limitations of static skill evolving solutions and formulate skill meta-evolving as a multi-level residual optimization problem. % for skill evolving paradigm.
    \item We propose HiSME, a lightweight text-based framework that uses unified execution feedback to improve both task-level skills and the evolving procedure that maintains them.
    \item We validate HiSME on several mainstream benchmarks through experiments and analyses, demonstrating the effectiveness of test-time meta-evolving in the evaluated settings.
\end{itemize}

\section{Preliminaries}

\label{sec:preliminaries}

\subsection{Agentic Systems}

Considering a ReAct-style agent system~\citep{react} based on LLMs with fixed parameters, we denote the \textbf{executor} as $\Exe$. Given a task $x$ from dataset $\mathcal{D}$, the executor generates a corresponding \textbf{trace} by taking actions and receiving observations until the task is solved in $N$ steps. This is regarded as a Markovian Decision Process (MDP) with initial state $x_0=x$:
\begin{equation}
\begin{aligned}
a_i&\sim\Exe(\cdot|x_i),\\
x_{i+1}&=\mathcal{T}(x_i, a_i, o_i)\coloneqq x_i\oplus\{(a_i, o_i)\},\\
y&\coloneqq x_N=\{x\}\cup\{(a_i, o_i)\}_{i=1}^N,
\end{aligned}
\end{equation}
denoted briefly $y\sim \Exe(\cdot|x)$. Its utility $U(y)$ is measured by task success, efficiency, or benchmark-specific scoring. 

\subsection{Skill Evolving as Approximate Parametric Update}
We aim to maximize
\begin{equation}
    J_0(\Exe) \coloneqq \mathbb{E}_{x\sim \mathcal{D},y\sim\Exe(\cdot|x)}\left[ U(y) \right].
\end{equation}

Since the executor is fixed, parametric update like $\Exe_{t+1}=
\Exe_t+\alpha_0\nabla J_0(\Exe_t)$ is infeasible. 
We instead propose \textbf{skills} $S\coloneqq\{s_i\}_{i=1}^{|S|}$, reusable artifacts distilled from past traces, conditioned on which the executor can solve future problems, 
\begin{equation}
    \Exe_S(\cdot|x)=\Exe(\cdot|x,S).
\end{equation}
In practice, we use a lightweight retriever to select relevant skills from the library and expose them to the executor. The design of skill retrieval is not the main focus of this paper. We elaborate our implementation details in appendix \ref{appendix:retrieval}.

Let $\widehat S_0$ denote the initial skill library, and in our setting $\widehat S_0=\emptyset$. 
To optimize the expected utility $J_0$, a \textbf{skill evolving} algorithm $\Evo$ consistently updates the skill library by learning from datasets $\mathcal{D}_t$ and corresponding traces generated by $\Exe$ at each timestep $t$,
\begin{equation}
\widehat S_{t+1}\sim \Evo(\widehat S_t,\mathcal D_t,\Exe), 1\le t\le T,
\end{equation}
% in order to achieve the optimal expected utility $J_0$.
\begin{equation}\label{eq:skill_evolving}
    \widehat S_T \approx \arg\max_S J_0(\Exe_S).
\end{equation}
From this view, skill evolving approximates the executor's ideal parametric update step.
\section{Methodology}

We first illustrate the high-level framework of our method in \S \ref{sec:theory}, then elaborate the design of each algorithmic role in \S \ref{sec:skill_generation} to \ref{sec:meta_maintenance}.

\begin{figure*}[t]
\centering
\includegraphics[width=\textwidth]{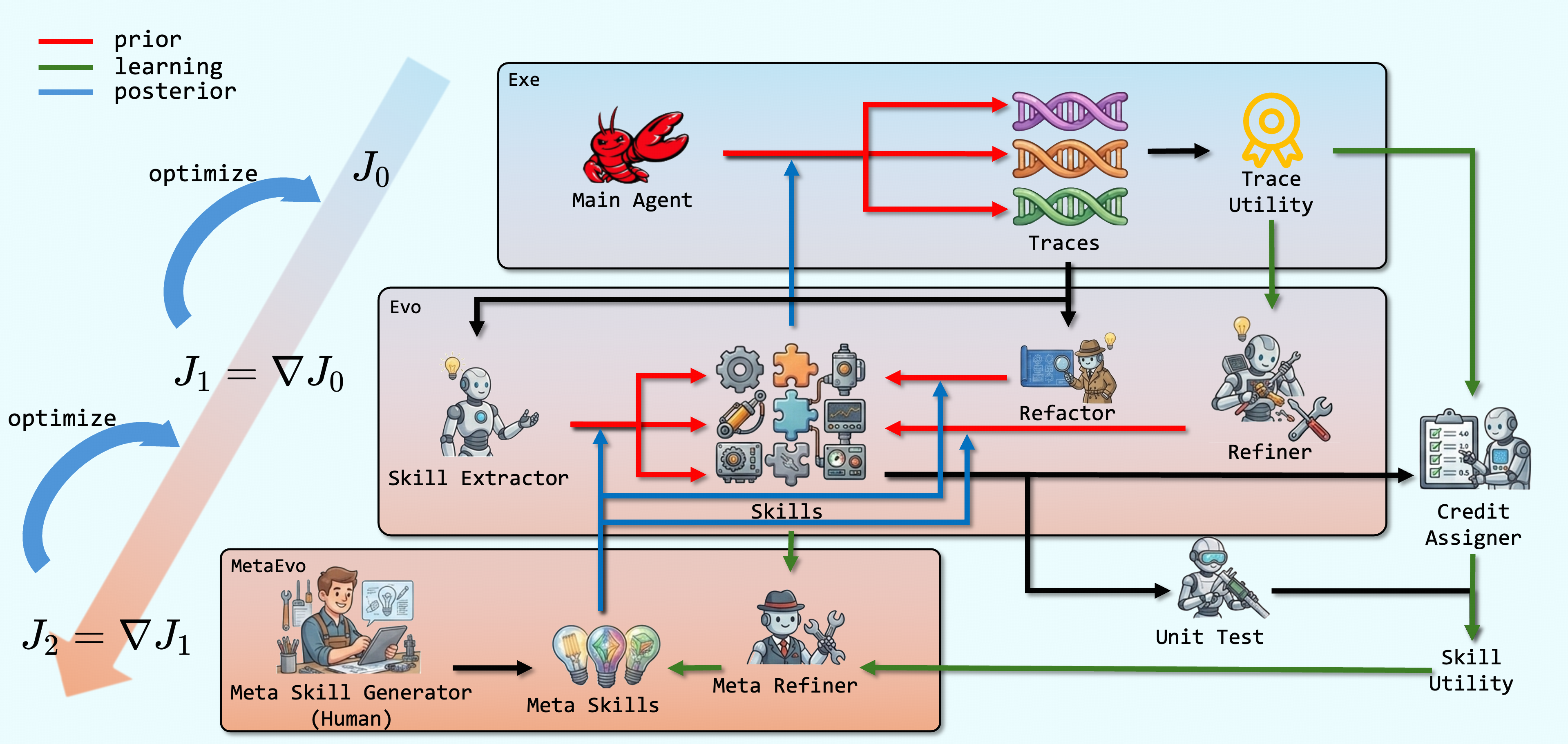}
\caption{Overview of HiSME. Red arrows denote pre-hoc guidance to the executor, blue arrows denote posterior evidence from traces to skills, and green arrows denote learned feedback to the evolving procedure.}
\label{fig:overview}
\end{figure*}

\subsection{Hierarchical Residual Optimization}
\label{sec:theory}

If we dive deeper into the perspective proposed in \S \ref{sec:preliminaries}: considering the agent system (executor) as a flawed prior for solving a problem:
\begin{equation}
    \epsilon_0=\max_\Exe J_0(\Exe)-J_0(\Exe_0)>0,
\end{equation}
\noindent A skill-evolving algorithm serves as a posterior optimizer on the \textbf{residual of its utility} at test time. The improved utility from a skill evolving algorithm $\Evo_0$ on the executor is
\begin{equation}
\begin{aligned}
J_1(\Evo_0)=
&\mathbb{E}_{\{S_t\}_{t=1}^{T}\sim
\Evo_0(S_0,\mathcal D,\Exe_0)}\\
&\left[J_0(\Exe_{S_T})-J_0(\Exe_{0})\right].
\end{aligned}
\end{equation}
\noindent Hopefully $J_1>0$ if the algorithm is valid.

As a natural extension of this perspective, we introduce the core claim of HiSME: \textbf{the skill evolving algorithm $\Evo$ itself, designed with human prior, can also be flawed, and thus further optimized at test time}:
\begin{equation}
    \epsilon_1=\max_\Evo J_1(\Evo)-J_1(\Evo_0)>0.
\end{equation}
The remedy is the same as how skill evolving works on the executor: we can learn run-time experiences, namely \textbf{meta-skills}, to improve the evolving algorithm as a posterior:
\begin{equation}
\widehat{\Evo}_{t+1}=
\widehat{\Evo}_t(\cdot|M_t),
\end{equation}
\noindent where $M_t$ denotes the meta-skill set at timestep $t$. HiSME implements a meta-evolving algorithm that periodically updates the meta-skills:
\begin{equation}
    M_{t+1}=\MetaEvo(M_t,\mathcal{D}_t).
\end{equation}

As summarized in Table~\ref{tab:hisme_order} from the perspective of optimization, skills represent the first-order information of trace utility, and meta-skills represent the second-order information of trace utility.
We elaborate in the following subsections on the detailed design of HiSME on skill and meta-skill evolving. The overall hierarchy is illustrated in Figure~\ref{fig:overview}.

\begin{table*}[t]
\centering
\small
\begin{tabular}{p{3.2cm}p{0.8cm}p{10cm}}
\toprule
Object & Order & Role in HiSME \\
\midrule
Trace & 0 & Ultimate utility target produced by the frozen executor. \\
Skill evolving & 1 & Create, revise or filter skills as experiences to optimize trace utility. \\
Skill meta-evolving & 2 & Create, revise or filter meta-skills as experiences to optimize skill utility. \\
\bottomrule
\end{tabular}
\caption{Optimization hierarchy in HiSME.}
\label{tab:hisme_order}
\end{table*}

\subsection{Skill Generation}
\label{sec:skill_generation}

HiSME generates skills from the execution traces and corresponding utility feedback with two strategies.
The \textbf{extractor} receives each single trace and identifies potential reusable fragments or experiences according to human engineer's prior knowledge embodied in its general ability. The \textbf{refactorer} receives groups of similar or related traces or skills and abstracts potential shared structures or knowledge. For refactoring, we design a two-level ranking pipeline: first build a graph with nodes being all the segments from historical traces, and edge weights being semantic similarity measured by embeddings and N-gram statistics. Then we heuristically extract clique from this graph to find potentially similar segments. The final top-k cliques will be passed to the refactorer for concise reusable skills. The details of overlap graph construction and maintenance are provided in Appendix~\ref{appendix:overlap_graph}.

The resulting new skills at step $t$ is
\begin{equation}
\mathcal{G}_t =
\Ext(\tau_t)
\cup
\Rfc(\mathcal{C}_t),
\end{equation}
where $\tau_t$ is the current trace, $\mathcal{C}_t$ is the set of overlap-supported trace/skill groups. 

Following~\citep{skillx}, we consider three types of skill semantics in this work: callable functions, workflow frameworks, and domain-specific knowledge. The full skill bodies are injected into context of the executor directly.

\subsection{Skill Evaluation and Maintenance}
\label{sec:skill_maintenance}

Since the skills are generated basically from LLMs' prior knowledge, they require further evaluation according to their future usage and maintenance based on the evaluation signals.

For evaluation, HiSME checks both short-term and long-term utility of a skill. In respect of short-term utility, the \textbf{bundle tester} builds and maintains with LLMs a test bundle for each skill, including both unit test cases and integration test cases. It mainly checks whether a skill implements the contract it declares faithfully. In respect of long-term utility, the \textbf{credit assigner} uses real executor traces to estimate with LLMs whether each exposed skill was helpful, harmful or neutral to the given task. Besides, \textbf{usage statistics} mechanism will record the lifetime events for each skill, among which here we concern retrieval and execution times.

Putting all these together, HiSME keeps an evidence state for each skill:
\begin{equation}
\label{eq:evidence_state}
e_t(s)=\{B(s),\ c_t(s),\ u_t(s)\}.
\end{equation}
Here $B(s)$ is the latest bundle for the current version of the skill. $c_t(s)$ are credit events collected after exposure or use. $u_t(s)$ are lifetime events.

With these evaluation signals, maintenance on the skill library becomes possible. HiSME adopts mainly two types of maintenance mechanism. First, the \textbf{refiner} revises with LLMs an existing skill when its evidence state indicates that its scope, interface, implementation, or trigger condition can be improved. The corresponding test bundle may alter as well due to the skill being revised:
\begin{equation}
\begin{aligned}
(s_{t+1}, B(s_{t+1})) &\sim \Rfn(s_t,e_t(s_t)),\\
\release(s_{t+1})&=\mathbbm{1}[\Gate(s_{t+1},B(s_{t+1}))].
\end{aligned}
\end{equation}
The refined version is released only after passing the maintained bundle. 

Second, \textbf{filter} mechanism simply changes the exposure level of a skill according to its performance. Skills with explicit harmful evidence and no helpful evidence can be marked as disabled. Let $\mathcal{E}_s$ denote the set of credit events assigned to skill $s$ in a
maintenance window. The filter first aggregates two counts:
\begin{align}
h_s&=\sum_{e\in\mathcal{E}_s}\mathbbm{1}[j_e=\mathrm{harmful}],\\
p_s&=\sum_{e\in\mathcal{E}_s}\mathbbm{1}[j_e=\mathrm{helpful}],
\end{align}
With the credit-filter threshold $\tau$ and the positive-protection threshold $\tau_p$, the filter gate is
\begin{equation}
\Filter(s) =
h_s\ge \tau\ \wedge\ p_s<\tau_p.
\end{equation}
\noindent Thus, a skill is disabled only when harmful evidence exceeds the filter threshold and the skill lacks enough positive evidence.

\subsection{Meta-skill Generation}
\label{sec:meta_maintenance}

Analogous to optimizing skills with evolving framework in \S \ref{sec:skill_maintenance}, the maintenance algorithm itself is not perfect and requires evaluation and optimization.

For meta-evaluation, HiSME stores experiences as meta-skills for each LLM-based algorithm $r\in\{\Ext,\Rfc,\Rfn\}$ in a role-specific replay buffer $\mathcal{B}^r_t$. An experience row for role $r$ links each produced artifact to its later utility:
\begin{equation}
\mathcal B^r_t=\{\rho^r(s)=(s, e_t(s)): s\text{ is produced by }r\},
\end{equation}
where $s$ is the skill or skill group produced from that input, and $e$ is the utility observed after publication or refinement, as defined in \eqref{eq:evidence_state}. 

For meta-evolving, HiSME periodically samples experience rows from the corresponding replay buffer of each role, and call an LLM agent $\MetaEvo$ to summarize one or more rules-of-thumb as meta-skills. In practice, we maintain no more than 5 pieces of meta-skills. Thus, we prompt $\MetaEvo$ to carefully decide which rules are most important to keep.
\begin{equation}
M^r_{t+1}=\MetaEvo^r(M^r_t,\Sample(\mathcal{B}^r_t)).
\end{equation}
The resulting meta-skills are supplied to the role's context as learned posterior:
\begin{equation}
r_t(\cdot)=r(\cdot;M^r_t),\quad r\in\{\Ext, \Rfc, \Rfn\}.
\end{equation}

In summary, HiSME is a simple but effective meta-evolving framework with the following advantages:
\begin{itemize}
    \item \textbf{Lightweight}. The only additional developing cost comes from adding a meta-skill extractor.
    \item \textbf{General}. The meta-evolving system of HiSME can be deployed to any LLM-based framework with context.
\end{itemize}
\section{Experiments and Analyses}

\subsection{Experimental Setup}

\textbf{Benchmark and dataset.} We evaluate HiSME on two agentic settings. BFCL-v4 multi-turn base is our primary completed benchmark~\citep{bfcl}. It evaluates function selection, argument binding, state tracking, and official multi-turn tool-call validity. MineDojo evaluates long-horizon embodied interaction in an open-world environment~\citep{minedojo}.

\textbf{Metrics and baselines.} For BFCL-v4, we report \texttt{official\_valid}, the official BFCL multi-turn validation rate~\citep{bfcl}, as the primary metric, and \texttt{avg\_score}, the average partial-credit score assigned by the BFCL evaluator. We further report call-level recall and precision to diagnose tool-call coverage and correctness, respectively. 
For MineDojo, we report \texttt{success\_rate} as the primary metric and use average environment steps and executor--environment interaction rounds as efficiency diagnostics.

To make maintenance overhead explicit, we report token cost separately for training-time evolving and held-out evaluation. \texttt{Train MTok} counts all LLM input/output tokens consumed during online skill evolving, including the executor and all maintenance roles. \texttt{Test MTok} counts the frozen evaluation cost on held-out tasks. \texttt{Total MTok} is the sum of both and serves as the overall maintenance budget.

We compare our method with \texttt{HiSME-static} ablation intended for showing performance advantage against a mostly aligned static evolving algorithm. Since crucial for our main conclusion, we set this comparison in our main results. For recent skill evolving baselines, we compare our method with SkillX~\citep{skillx}, a recent related static evolving algorithm for skill lifetime maintenance. We re-implement their method based on their open-source implementation.\footnote{Appendix~\ref{app:skillx_executor_alignment} reports an executor-alignment check on BFCL-v4. Injecting the same SkillX library through our executor performed worse than the SkillX official-style runner, so the main table reports the stronger SkillX result.} For other test-time optimization methods, we compare AWM~\citep{awm} and MementoSkills~\citep{mementoskills}. AWM is a memory-based agent that induces reusable workflows from successful trajectories and retrieves them at test time to guide problem solving. MementoSkills follows a similar philosophy but distills reflective knowledge skills from prior experience.

\textbf{Training configuration.} All experiments use \texttt{claude-sonnet-4-6} as both the base ReAct executor and the LLM behind maintenance roles. We use deterministic decoding for executor/tool-use evaluation. For BFCL-v4, the executor outputs thoughts and tool callings and observes tool call feedbacks. For MineDojo, the executor outputs Python programs representing primitive compositions to act in the environment, and observes the resulting environment state. The max rounds for these benchmarks are 20 and 100, respectively.

For each benchmark, we use a fixed train/test split: training tasks are used for online skill evolving, while held-out test tasks are evaluated with a frozen skill repository. The sizes of training set / test set for these benchmarks are 150 / 50 and 30 / 30, respectively. Three random runs are conducted on each method. For our methods, we invoke refactoring, filtering and meta-skill updating at every 5 tasks. Our method design is based on single-thread assumption. For efficiency concern, however, in practice we run all tasks between two meta updates in parallel and update skills following a thread-safe mechanism.

We sample 3 extractor candidates for each input because single-trace extraction has higher uncertainty and is more sensitive to prompt priors. Both refactorer and refiner generates one candidate at each time. Implementation details of prompt formats, truncation limits, skill fields, bundle fields, retry budgets, graph parameters, and retrieval score constraints are elaborated in Appendix \ref{appendix:retrieval}.

\subsection{Main Results}

\begin{table*}[t]
\centering
\footnotesize

\textbf{(a) BFCL-v4}

\vspace{0.25em}

\tabresize{
\begin{tabular}{lrrrrr}
\toprule
Method & Official Valid Rate $\uparrow$ & Avg Score $\uparrow$ & Train MTok & Test MTok & Total MTok \\
\midrule
NoSkill       & $\text{0.580}_{\pm\text{0.035}}$ & $\text{0.778}_{\pm\text{0.013}}$ & 0.00  & 3.37 & 3.37 \\
SkillX        & $\text{0.747}_{\pm\text{0.046}}$ & $\text{0.782}_{\pm\text{0.007}}$ & 10.20 & 4.21 & 14.41 \\
AWM           & $\text{0.647}_{\pm\text{0.012}}$ & $\text{0.797}_{\pm\text{0.009}}$ & 10.23 & 3.62 & \textbf{13.85} \\
Memento       & $\text{0.607}_{\pm\text{0.042}}$ & $\text{0.786}_{\pm\text{0.011}}$ & 11.25 & 3.71 & 14.96 \\
HiSME-static  & $\text{0.760}_{\pm\text{0.020}}$ & $\text{0.820}_{\pm\text{0.003}}$ & 11.35 & 3.68 & 15.03 \\
HiSME         & $\textbf{0.800}_{\pm\text{0.053}}$ & $\textbf{0.827}_{\pm\text{0.011}}$ & 11.50 & 3.87 & 15.37 \\
\bottomrule
\end{tabular}}

\vspace{0.9em}

\textbf{(b) MineDojo}

\vspace{0.25em}

\tabresize{
\begin{tabular}{lcccccc}
\toprule
Method & Success Rate $\uparrow$ & env\_steps $\downarrow$ & exec\_rounds $\downarrow$ & Train MTok & Test MTok & Total MTok \\
\midrule
No Skill      & $\text{0.689}_{\pm\text{0.107}}$ & $\text{10224.0}_{\pm\text{397.4}}$ & $\text{46.8}_{\pm\text{5.8}}$ & 13.80 & 13.55 & 27.35 \\
SkillX        & $\text{0.733}_{\pm\text{0.033}}$ & $\text{12008.3}_{\pm\text{1849.6}}$ & $\text{60.8}_{\pm\text{6.9}}$ & 19.47 & 14.34 & 33.81 \\
AWM           & $\text{0.567}_{\pm\text{0.088}}$ & $\textbf{6060.0}_{\pm\text{1098.5}}$ & $\text{60.0}_{\pm\text{7.5}}$ & 17.37 & 16.73 & 34.10 \\
Memento       & $\text{0.833}_{\pm\text{0.067}}$ & $\text{7489.6}_{\pm\text{520.4}}$ & $\text{42.7}_{\pm\text{5.6}}$ & 14.86 & 14.12 & 28.98 \\
HiSME-static  & $\text{0.700}_{\pm\text{0.033}}$ & $\text{8144.9}_{\pm\text{2104.3}}$ & $\text{62.9}_{\pm\text{4.8}}$ & 20.68 & 18.99 & 39.67 \\
HiSME         & $\textbf{0.856}_{\pm\text{0.019}}$ & $\text{6984.7}_{\pm\text{1227.8}}$ & $\textbf{40.3}_{\pm\text{2.4}}$ & 14.42 & 13.41 & \textbf{27.83} \\
\bottomrule
\end{tabular}}
\caption{Held-out test results on BFCL-v4 (top) and MineDojo (bottom). Arrows indicate the preferred direction. Bold highlights the best value among skill-based methods in each column; No Skill is included as a reference baseline.}
\label{tab:main_results}
\end{table*}

\begin{table*}[t]
\centering
\footnotesize
\tabresize{
\begin{tabular}{lrrrrrr}
\toprule
Method & official\_valid $\uparrow$ & $\Delta$ & avg\_score $\uparrow$ & $\Delta$ & call recall $\uparrow$ & call precision $\uparrow$ \\
\midrule
HiSME & \textbf{0.7800} & - & \textbf{0.8511} & - & \textbf{0.9040} & \textbf{0.8136}  \\
\midrule
\ \ static & 0.6939 & -0.0861 & 0.8143 & -0.0368 & 0.8670 & 0.7775  \\
\ \ w/o extractor & 0.6800 & -0.1000 & 0.8001 & -0.0510 & 0.8755 & 0.7501  \\
\ \ w/o refiner & 0.6939 & -0.0861 & 0.8046 & -0.0465 & 0.8730 & 0.7548  \\
\ \ w/o refactor & 0.5400 & -0.2400 & 0.7796 & -0.0715 & 0.8581 & 0.7339  \\
\ \ w/o credit filter & 0.7347 & -0.0453 & 0.8135 & -0.0376 & 0.8850 & 0.7647  \\
\bottomrule
\end{tabular}}
\caption{BFCL-v4 component ablations on the held-out test set.}
\label{tab:bfcl_ablation}
\end{table*}

Table~\ref{tab:main_results} summarizes the held-out results on BFCL-v4 and MineDojo. We highlight three main observations from these results.

\noindent\textbf{Skills act as effective test-time residuals.} Across all benchmarks, HiSME improves the frozen executor without changing model parameters. On BFCL-v4, the gain appears not only in official validity but also in call recall and precision, indicating that the repository changes the structure of tool-call trajectories rather than merely adding more prompt context. The increased token cost is due to producing more otherwise missing calls. On MineDojo, HiSME obtains the highest success rate while reducing execution rounds and total token use, suggesting that useful skills can shorten interaction even in long-horizon embodied tasks.

\noindent\textbf{Our framework has architectural advantage over the representative baseline work.} Our results show that HiSME-static already improves over SkillX, isolating the value of HiSME's static lifecycle. 

\noindent\textbf{Meta-evolving succeeds in optimizing further residual utility of the agent system.} HiSME-static already provides a strong static evolving framework through bundle-gated release, posterior refactoring, credit-based repair, and conservative filtering. Full HiSME uses the same framework as the static variant except for adding a meta-evolving mechanism and substantially improves over it. The performance gain is therefore attributable to learned role guidance that affects how candidate skills should be built and maintained.

In the following sections, we perform further analyses on BFCL-v4 benchmark in consideration of cost.

\subsection{Ablation Studies}

We ablate the key algorithmic roles of HiSME to explore the effect of each role on the final performance. The results are in Table \ref{tab:bfcl_ablation}. 

\noindent\textbf{Reusable structure requires cross-trace evidence.} We find that removing refactor causes the largest degradation, making it the most important skill generation component in BFCL-v4. This supports the role of the overlap graph: reusable procedures and experiences are more reliably identified from shared evidence than from isolated traces.

\noindent\textbf{Local proposal and posterior consolidation are complementary.} Removing the extractor also hurts performance, even with refactoring enabled. Despite with least evidence, the human engineer's prior on reusability can be a strong source of performance gain complementing cross-trace evidence from real use cases. 

\noindent\textbf{Post-hoc evaluation and maintenance complete the closed-loop skill generation procedure. } Removing either the refiner or the filter mechanism shows considerable degradation on performance. This suggests that generated skills require further test and maintenance to function well in future use. The refiner shows higher utility against the filter due to enriched operations on skill semantics, but filtering out harmful skills serves as a important role as well.

\subsection{Process Evaluation}

We further study the dynamic during evolving of HiSME and how it differs from static algorithm. Specifically, we save and analysis the intermediate skill libraries every fixed steps. We first test their performance on held-out test set. Then we perform credit assignment with the same pipeline during training on these skills to see their specific contribution to the final results.

\begin{figure*}[htbp]
    \centering
    \begin{minipage}[t]{0.48\textwidth}
        \centering
        \includegraphics[width=\linewidth]{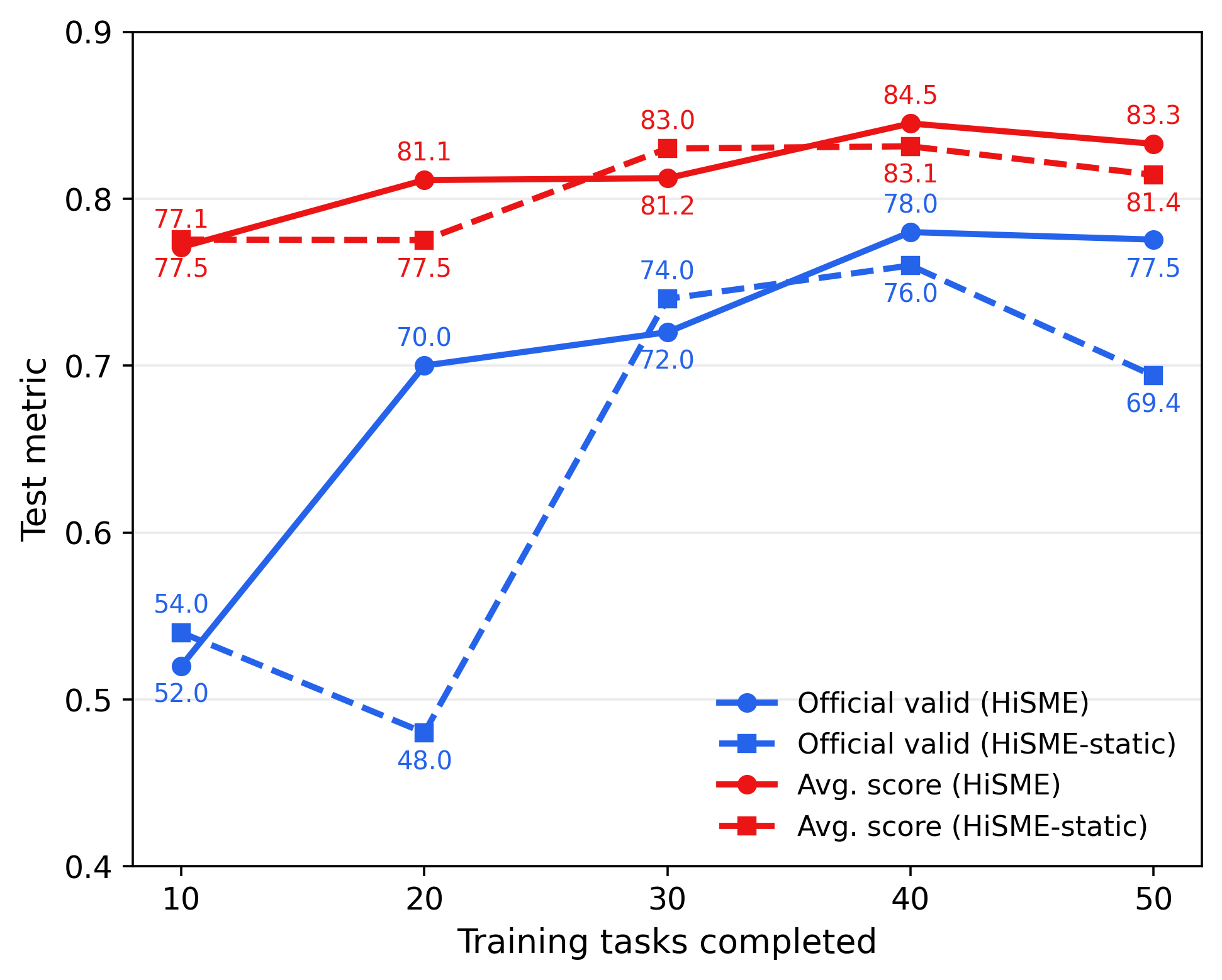}
        \caption*{(a)}
        \label{fig:metric}
    \end{minipage}
    \hfill
    \begin{minipage}[t]{0.42\textwidth}
        \centering
        \includegraphics[width=\linewidth]{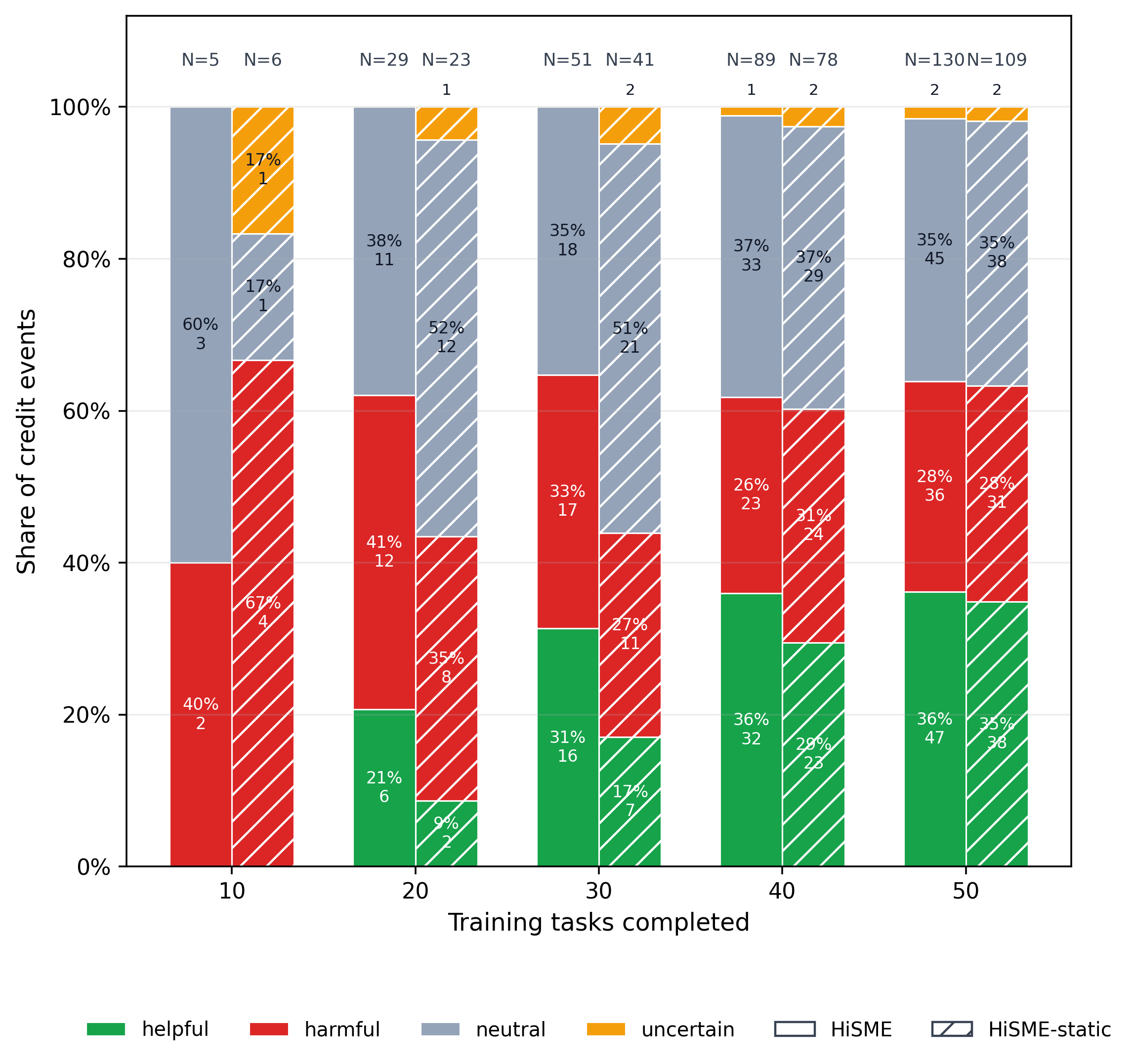}
        \caption*{(b)}
        \label{fig:credit}
    \end{minipage}
    \caption{Process evaluation results compared between HiSME and HiSME-static. (a) Test result with skills accumulated during training. (b) Analyses on credits of each skills on test time in four categories. }
    \label{fig:process}
\end{figure*}

As shown in Figure \ref{fig:process}(a), both of our evolving algorithms consistently improves test performance as evolving proceeds. Meanwhile, the test performances of meta-evolving overall surpass skill evolving during training. This suggests that evolving on skill evolving algorithm itself can improve the skill quality during most of the training process. In Figure \ref{fig:process}(b), both algorithms produce increasing helpful skills and decreasing harmful skills, while meta-evolving again surpasses static skill evolving by enhancing more and harming less performances of the executor.

\subsection{Meta Test}

Results above suggest that HiSME generates better skill library compared with static algorithm. We further explore whether the produced meta-skills bring transferable experience for future skill evolving. We take the role-level meta-skills learned by a completed HiSME run, freeze them, and use them as initialization for a new static evolving run. All other settings are aligned with cold-start HiSME-static. This setting tests whether the learned evolving rules can improve a future evolving process.

The results are shown in Table~\ref{tab:meta-test}. Frozen meta-skills substantially improve \texttt{official\_valid} over cold-start static evolving, nearly matching full HiSME on this metric, while maintaining close \texttt{avg\_score}. This indicates strong but not full transfer: the learned role guidance helps the next static run produce more officially valid complete traces, but does not fully preserve the fine-grained tool-call quality of cold-start static evolving.

\begin{table*}[t]
\centering
\footnotesize
\tabresize{
\begin{tabular}{lrrrrrrr}
\toprule
Method & official\_valid $\uparrow$ & avg\_score $\uparrow$ & call recall $\uparrow$ & call precision $\uparrow$ & total tok.  $\downarrow$ \\
\midrule
static from scratch & 0.6939 & \textbf{0.8143} & 0.8670 & \textbf{0.7775} & \textbf{69865.7} \\
static from meta & \textbf{0.7708} & 0.8020  & \textbf{0.8847} & 0.7440 & 70805.1 \\
\bottomrule
\end{tabular}}
\caption{BFCL-v4 meta-test results. Frozen-meta static uses meta-skills learned by HiSME as fixed role guidance for a new static evolving run, while disabling online meta-skill updates. Deltas are computed against cold-start static evolving.}
\label{tab:meta-test}
\end{table*}

\subsection{Case Studies}
\label{sec:case_studies}

We include two qualitative studies to inspect whether HiSME's improvements come from grounded maintenance rather than accidental prompt expansion. Appendix~\ref{app:case_studies} gives the detailed evidence.

\noindent\textbf{Meta-skill evolution is grounded in observed candidate behavior.} We trace five learned extractor meta-skills back to concrete evidence from BFCL-v4 candidate groups. The learned rules do not merely restate generic prompting advice. They distinguish schema-grounded contracts and cross-turn state dependencies from brittle rules triggered by surface words such as ``urgent'' or by stylistic preferences such as message brevity. This supports our interpretation that meta-evolving learns operational constraints for future skill generation from real skill outcomes.

\noindent\textbf{Skill version updates make targeted semantic changes.} We examines how skills are actually revised. Here are two representative skills: a TradingBot order-confirmation workflow and a VehicleControlAPI brake-before-start rule. In both cases, HiSME keeps the useful invariant but changes the condition under which it should apply: one update narrows a workflow trigger to immediately preceding order placement, while the other specifies the exact brake-pedal argument required by the API. This shows that skill evolution is not unrestricted rewriting; it uses task evidence to make local, testable changes to scope or contract.

% Rebuttal results

\section{Related Work}

\subsection{Static and RL-Based Skill Evolving}

\noindent\textbf{Trace-to-skill extraction.} A first line of work studies how to convert agent experience into reusable external skills. Trace2Skill distills trajectory-local lessons into transferable skills~\citep{trace2skill}, while SkillX organizes planning knowledge, functional skills, and atomic skills into a multi-level skill base~\citep{skillx}. Related systems such as SkillForge, SkillGen, and SkillMaster further explore memory-like skill stores, domain-specific support skills, inference-time skill synthesis, and autonomous skill editing~\citep{skillforge,skillgen,skillmaster}.

\noindent\textbf{Skill repositories and lifecycle control.} A second line focuses on how skills should be stored, verified, selected, and maintained after creation. CoEvoSkills and SkillClaw emphasize verifier-guided or repository-level skill evolving~\citep{coevo_skills,skillclaw}. SkillLens studies hierarchical and multi-granularity reuse~\citep{skilllens}. SkillMOO optimizes skill bundles under multiple objectives such as pass rate and cost~\citep{skillmoo}, while PSN and SkillsVote study maturity-aware updates, rollback, recommendation, and evidence-gated lifecycle governance~\citep{psn,skillsvote}. These works show that useful skills require long-term repository management, not merely one-shot extraction.

\noindent\textbf{RL-based and multi-agent skill adaptation.} A third line couples skills with reinforcement learning or multi-agent adaptation. SkillRL, Skill1, D2Skill, XSkill, SkillMAS, and OpenClaw-RL use downstream rewards, verbal feedback, or agent restructuring to improve skill selection, utilization, distillation, or multi-agent behavior~\citep{skillrl,skill1,d2skill,xskill,skillmas,openclaw_rl}. These methods demonstrate that skill-related adaptation can be optimized from task outcomes, although the learned executor, policy, or multi-agent controller is usually the central object of optimization.

HiSME is related to all three directions. It also extracts and maintains skills, but focuses on using later skill outcomes to guide the runtime evolving procedure itself.

\subsection{Text-Based Optimization}

\noindent\textbf{Prompt-level search.} OPRO formulates optimization as a natural-language search process in which the language model proposes new candidate prompts or solutions from previous trials~\citep{opro}. This line shows that textual hypotheses can be iteratively improved without gradient updates to model parameters.

\noindent\textbf{Textual gradients and critiques.} ProTeGi introduces natural-language ``gradients'' that describe how a prompt should change~\citep{protegi}, and TextGrad generalizes this idea into a framework where textual feedback plays a role analogous to differentiation~\citep{textgrad}. These methods use language feedback to revise textual objects rather than dense parameters.

\noindent\textbf{Agent reflection and verbal reinforcement.} Reflexion uses verbal feedback from prior trajectories to improve later agent behavior~\citep{reflexion}. Similar reflection-style methods treat natural-language memory as a compact carrier of past failure modes and repair strategies.

HiSME adopts the same broad premise that text can be an optimization medium. Its optimized object, however, is a persistent skill repository and the role-level policy that creates, revises, and filters skills.

\section{Conclusion}

We introduce HiSME, a lightweight and efficient skill meta-evolving framework. By modeling skill evolving as a multi-level higher-order optimization problem over the utility of the final execution outcome, the skill and skill maintenance algorithms themselves can be automatically and adaptively tuned. Meta-skills are shown to be a simple and effective adaptive optimization method for LLM agent algorithms. The experiments not only demonstrate the advantages of the skills produced by our algorithm, but also reflect the advantages of meta-skills in downstream task adaptation.

\section*{Limitations}

\textit{Inaccurate credit } The framework still depends on LLM-based credit assignment, bundle maintenance, and refinement. When those judgments are wrong, useful skills can be over-pruned or harmful skills can survive longer than they should.

\noindent\textit{Extra train cost } The system introduces extra runtime cost because maintenance and meta-maintenance require additional model calls. This cost is acceptable when skills are reused many times, but it is a real tradeoff in short-lived or low-reuse settings.

\noindent\textit{Limited skill format } We only consider rules-of-thumb as meta-skills. More enriched formats, like callable functions and workflows, are possibly beneficial for meta evolving. Furthermore, higher levels of evolution are also possible for automatically optimizing meta-evolving framework.  We leave these possibilities for future works to explore.

\bibliography{references}

\clearpage
\appendix
\section{Benchmark Details}

\subsection{BFCL-v4: Function Calling and Multi-Turn Tool Use}

Berkeley Function Calling Leaderboard (BFCL) was introduced by the UC Berkeley Gorilla team as a benchmark for evaluating whether language models can select and invoke external functions correctly~\citep{bfcl}. The official BFCL description frames function calling as the ability to map a natural language request into one or more valid API/tool invocations, including correct function names, argument schemas, values, ordering, and cases where no function should be called. The benchmark covers serial and parallel function calls, multiple programming languages, AST-based evaluation, executable evaluation, and later versions add more realistic multi-turn and multi-step agentic settings.

In this paper, we use BFCL-v4 as the primary benchmark for online skill evolving because it stresses exactly the type of behavior a skill repository is supposed to improve: repeated argument binding, ordering-sensitive workflows, schema obedience, state tracking across turns, and tool-use recovery after errors. A BFCL task is not simply a question-answer pair. It provides a conversation, a tool environment, and an official expected call trace. The model must produce calls that are both semantically appropriate and compatible with the tool schema. This makes BFCL a useful setting for measuring whether a skill helps the executor call tools more reliably, or whether it pollutes the prompt and causes extra or malformed calls.

Our local BFCL related-task split uses a curated subset of BFCL-v4 multi-turn tasks. The current recommended setting is a deterministic 150/50 split: 150 training tasks for online evolving and 50 held-out tasks for frozen-store evaluation. Training is serial because the skill repository, overlap graph, and feedback memory are mutated after each task. Held-out evaluation is read-only and can run concurrently. The related-task manifest ranks examples by repeated tool-family patterns, lookup/action mixtures, path length, and multi-turn argument-binding failure modes.

One representative BFCL training task is \texttt{multi\_turn\_base\_118}, a TradingBot workflow. The user first asks to inspect a stock watchlist, then remove the first stock, then buy 100 shares of AAPL at the current market price, then inspect the latest order, and finally cancel that order. The expected trace is:

\begin{verbatim}
turn 1: get_watchlist()
turn 2: remove_stock_from_watchlist(symbol='NVDA')
turn 3: get_stock_info(symbol='AAPL')
        place_order(order_type='Buy', symbol='AAPL', price=227.16, amount=100)
turn 4: get_order_details(order_id=12446)
turn 5: cancel_order(order_id=12446)
\end{verbatim}

This example tests several reusable behaviors. The executor must bind the watchlist item \texttt{NVDA} from the first call into the remove operation; it must lookup the current AAPL price before placing a market-price order; and it must carry the returned order id into later inspection and cancellation calls. A useful skill for this region is not "always buy AAPL", but a narrower workflow rule: when a trading request asks for current-market execution, first retrieve the stock information, then place the order using the observed price, and preserve the generated order identifier for subsequent order-management turns. This is why our credit assigner creates focused bundle fragments around the relevant tool-call subsequence instead of turning the entire conversation into a single coarse regression case.

Another held-out example, \texttt{multi\_turn\_base\_51}, combines two tool families: vehicle control and messaging. The expected trace locks all doors, starts the engine only after pressing the brake pedal, checks tire pressure and finds a nearby tire shop, then logs into the message API and sends a specific message from \texttt{USR001} to \texttt{USR002}. This tests cross-domain tool selection and precondition ordering. A skill that teaches "press brake before startEngine" can help the vehicle subtask, but it should not influence the later message API turn. This motivates our attribution-scope field in credit assignment and our dependency-aware refiner: a local vehicle workflow should be refined or tested on the vehicle fragment, not on the unrelated messaging fragment.

\subsection{MineDojo: Long-Horizon Embodied Control in Open-World Minecraft}

MineDojo is an embodied-agent benchmark built on Minecraft for open-ended, language-conditioned interaction in procedurally generated 3D worlds~\citep{minedojo}. We use MineDojo to evaluate long-horizon embodied resource collection and crafting. Unlike BFCL-v4, where the main challenge is to produce valid symbolic tool calls, MineDojo requires the executor to interact with a live environment over many rounds: it must observe the local world state, move and rotate the camera, mine or attack blocks, collect dropped items, craft intermediate objects, equip tools, and recover from failed low-level actions.

Our setting differs from several Minecraft-based agent benchmarks and systems that expose higher-level Mineflayer-style JavaScript APIs. Such interfaces often provide semantically meaningful helper functions for navigation, block collection, or crafting, which can hide much of the low-level control problem. In contrast, our MineDojo executor operates through the environment's atomic Minecraft action space, including movement, camera control, attack, use, craft, equip, place, and destroy. Thus, a successful trajectory must compose primitive grounded actions rather than call a prebuilt semantic routine such as collecting a block or navigating to an object.

We also do not use a visual language model in the MineDojo experiments. Although MineDojo can provide egocentric RGB observations, our evaluation uses a purely text-based interaction interface. To make this possible, we modify the environment interface to expose MineDojo's ray-based perception signal. The ray channel reports visible blocks and entities in the agent's field of view. This gives the LLM executor grounded local affordances without relying on image recognition or providing a global map.

Our local MineDojo split focuses on programmatic Harvest and Tech Tree tasks, where success can be measured from the final inventory or item-use state. We use 30 training cases for online evolving and 30 held-out test cases for frozen-store evaluation. Both splits are organized into batches of five cases.

We use a dynamic CodeAct-style executor for MineDojo. At each round, the client converts the current environment state into a textual prompt containing the task, previous execution result, cached Minecraft knowledge queries, and compact ray-based world state. The LLM then either issues an auxiliary knowledge query, such as a recipe or item lookup, or writes executable Python control code. The code defines a \texttt{run(controller, obs)} function that operates through MineDojo's primitive controller actions. After execution, the resulting logs, errors, inventory changes, and updated observation are returned to the next round, allowing the LLM to iteratively repair failed attempts and continue the embodied task. Because this interface uses primitive MineDojo actions and text-only ray observations, it is not directly comparable to Minecraft agents that rely on Mineflayer-style semantic APIs or visual-language perception.

\section{Retrieval Implementation Details}

\label{appendix:retrieval}
Retrieval is used only to expose a small number of relevant active skills to the frozen executor. For each turn, HiSME builds a query from the current user request, the dialogue state, recent tool errors, and the previous assistant message. Each active skill is represented by a compact retrieval view containing its name, description, trigger conditions, allowed tools, domains, and a short body summary. We score each skill by a mixture of sparse lexical overlap, semantic similarity, tool/domain compatibility, and lifecycle trust:
\begin{equation}
\begin{aligned}
\rho(q,s)=&
\lambda_1\rho_{\mathrm{sparse}}(q,s)+
\lambda_2\rho_{\mathrm{emb}}(q,s)\\
&+\lambda_3\rho_{\mathrm{tool}}(q,s)+
\lambda_4\rho_{\mathrm{trust}}(s).
\end{aligned}
\end{equation}
Only active skills are exposed during held-out evaluation. Trial skills may be exposed during training with lower priority, but archived and disabled skills are never retrieved. The top-$K$ skills are rendered as prompt-only context in BFCL; they are not converted into callable tools in this setting.

\section{Refactorer Implementation based on Overlap-Graph}
\label{appendix:overlap_graph}

Overlap-graph refactoring is used to discover reusable structure that is weakly supported by a single trace but stable across related traces or existing skills. HiSME constructs a graph whose nodes are projected trace segments and skill summaries. A segment contains the task fragment, relevant tool calls, errors, state transitions, and outcome. For each node pair $(u,v)$, the edge weight combines sparse n-gram overlap, embedding similarity, and error-text overlap:
\begin{equation}
\begin{aligned}
w(u,v)=&
\alpha\,\mathrm{sparse}(u,v)+
\beta\,\mathrm{emb}(u,v)\\
&+\gamma\,\mathrm{err}(u,v).
\end{aligned}
\end{equation}
Edges below threshold are removed. We then search for small strict cliques with at least two distinct source tasks unless the group is explicitly a skill-revision group. Each accepted group must have an explainable purity signal: shared tools, shared precondition failure, shared argument-binding pattern, or semantically aligned task structure. The refactorer receives only these high-purity groups, which reduces the risk of merging traces that merely share superficial vocabulary.

The numerical graph parameters are benchmark-level implementation choices rather than the source of the main result. For the BFCL runs reported in this draft, we use $\alpha=0.45$, $\beta=0.35$, $\gamma=0.20$, an internal error-overlap weight of $1.7$, and a minimum edge weight of $0.18$. In our experiments, small changes to these values do not alter the qualitative conclusion; they mainly affect which candidate groups are sent to the refactorer.

\section{SkillX Executor Alignment on BFCL}
\label{app:skillx_executor_alignment}

For the BFCL comparison, we additionally evaluated whether SkillX should be reported under our executor for stricter runtime alignment. This variant reuses the same SkillX library and retrieval context but injects it through our BFCL executor. As shown in Table~\ref{tab:skillx_executor_alignment}, this executor-aligned variant performs worse than the SkillX official-style runner on official validity and average score. We therefore report the stronger SkillX official-style result in the main table, which is conservative with respect to our method.

\begin{table*}[t]
\centering
\footnotesize
\begin{tabular}{lccccc}
\toprule
SkillX variant & official\_valid $\uparrow$ & avg\_score $\uparrow$ & recall $\uparrow$ & precision $\uparrow$ & total tok. $\downarrow$ \\
\midrule
Official-style runner & 0.6600 & 0.7689 & 0.8837 & 0.7026 & 74664.9 \\
Through our executor & 0.5000 & 0.7545 & 0.8364 & 0.7007 & 68850.4 \\
\bottomrule
\end{tabular}
\caption{BFCL executor-alignment check for SkillX. Both rows use the same 50/50 shuffled split and SkillX library; the main table reports the stronger official-style SkillX result.}
\label{tab:skillx_executor_alignment}
\end{table*}

\section{Case Studies}
\label{app:case_studies}

We provide the source evidence behind the qualitative studies discussed in Section~\ref{sec:case_studies}. The first study inspects five learned extractor meta-skills. The second study follows one concrete skill across successive maintenance snapshots.

\subsection{Meta-skill Evidence}

Table~\ref{tab:meta_skill_cases} lists five extractor meta-skills and the evidence that led to them. The evidence comes from candidate-group feedback in the BFCL run \path{bfcl_meta_prompt_balanced_trl_50_50_20260523}. These cases show that the learned role guidance is grounded in observed helpful and harmful skill outcomes.

\begin{table*}[t]
\centering
\footnotesize
\begin{tabularx}{\textwidth}{>{\RaggedRight\arraybackslash}p{0.38\textwidth}>{\RaggedRight\arraybackslash}X}
\toprule
Meta-skill & Supporting evidence \\
\midrule
Extract parameter contracts only when multiple independent traces confirm the pattern; reject stylistic preferences unless they encode API requirements or prevent errors.
& An invoice-parameter skill received positive credit when the model omitted an unnecessary \texttt{insurance\_id}. In contrast, a support-message brevity skill received harmful credit because its guidance likely blocked a required support call for a vague user request. \\
\addlinespace
Do not extract workflow guardrails triggered by sentiment keywords or question patterns; require verifiable state, API preconditions, or cross-turn data dependencies.
& Travel support candidates triggered by words such as ``urgent'' or ``problem'' produced harmful or unsupported behavior. The feedback indicated that surface wording alone did not define a reusable workflow condition. \\
\addlinespace
Extract workflow guardrails for API preconditions only when at least two independent traces confirm the precondition and show successful application.
& Later candidate groups showed that single-trace API-precondition workflows could be context-sensitive. The extractor update therefore required repeated positive evidence before turning such observations into reusable rules. \\
\addlinespace
Do not extract skills that skip verification or status-check steps unless multiple positive traces confirm the optimization is safe.
& Vehicle and travel candidates that encouraged skipping checks caused extra calls or wrong arguments when the skipped state was needed to compute a valid action. The learned rule preserves verification unless evidence shows it is safely redundant. \\
\addlinespace
Merge workflow guardrail candidates that describe identical precondition sequences even if phrasing differs slightly; retain the clearer version.
& Candidate groups repeatedly contained near-duplicate variants, such as multiple invoice-omission skills or multiple brake-before-start vehicle skills. Competition and credit identified that these variants encoded the same underlying contract. \\
\bottomrule
\end{tabularx}
\caption{Examples of learned extractor meta-skills and their grounding evidence.}
\label{tab:meta_skill_cases}
\end{table*}

\subsection{Skill Evolution Trace}

Tables~\ref{tab:skill_evolution_trading} and~\ref{tab:skill_evolution_vehicle} show two cases where the skill body itself changes. They illustrate two common update types: \textit{narrow trigger} updates that restrict when a workflow applies, and \textit{specify argument} updates that make an API contract exact.

\begin{table*}[t]
\centering
\footnotesize
\begin{tabularx}{\textwidth}{>{\RaggedRight\arraybackslash}p{0.08\textwidth}>{\RaggedRight\arraybackslash}p{0.17\textwidth}>{\RaggedRight\arraybackslash}p{0.35\textwidth}>{\RaggedRight\arraybackslash}X}
\toprule
Version & Update & Explanation & Triggering event or update rationale \\
\midrule
v1
& New refactor skill.
& \path{trading_order_confirmation_workflow}: after \texttt{place\_order} returns an \texttt{order\_id}, if the immediately following turn asks to retrieve, check, or confirm order details without a new id, call \texttt{get\_order\_details} with the resolved id.
& Refactoring found repeated TradingBot trajectories where an order id produced by \texttt{place\_order} must be reused in the next turn. Positive evidence includes \texttt{multi\_turn\_base\_118}, where the workflow produced the correct order-detail lookup. \\
\addlinespace
v2
& Narrow trigger.
& The revised body makes the trigger explicit: apply only when \texttt{place\_order} was called in the immediately prior turn of the current task session and no intervening turn occurred. Do not apply merely because the user mentions a past order.
& Harmful evidence from \texttt{multi\_turn\_base\_143} showed workflow pollution: the task asked about a remembered past order without a prior \texttt{place\_order} in the current session. The skill biased the model toward an inapplicable confirmation workflow, so the refiner narrowed the precondition. \\
\bottomrule
\end{tabularx}
\caption{Skill-body update for \texttt{trading\_order\_confirmation\_workflow}.}
\label{tab:skill_evolution_trading}
\end{table*}

\begin{table*}[t]
\centering
\footnotesize
\begin{tabularx}{\textwidth}{>{\RaggedRight\arraybackslash}p{0.08\textwidth}>{\RaggedRight\arraybackslash}p{0.17\textwidth}>{\RaggedRight\arraybackslash}p{0.35\textwidth}>{\RaggedRight\arraybackslash}X}
\toprule
Version & Update & Explanation & Triggering event or update rationale \\
\midrule
v1
& New extractor skill.
& \path{vehicle_brake_before_engine_start}: before \texttt{startEngine(ignitionMode='START')}, call \texttt{pressBrakePedal}. The initial rule captures the brake-before-start ordering but does not emphasize the exact pedal position.
& Extracted from VehicleControlAPI evidence where engine start requires brake engagement before ignition. Subsequent positive credit showed the workflow aligned with correct engine-start traces. \\
\addlinespace
v2
& Specify argument.
& The revised body adds the exact contract: \texttt{pedalPosition} must be exactly \texttt{1.0}, and the rule should not apply if the brake is already confirmed pressed at \texttt{1.0}.
& A maintenance test showed the model could press the brake only partially, using \texttt{pedalPosition=0.5}. The refiner therefore changed the skill body from an ordering rule into an exact argument contract for the brake step. \\
\bottomrule
\end{tabularx}
\caption{Skill-body update for \texttt{vehicle\_brake\_before\_engine\_start}.}
\label{tab:skill_evolution_vehicle}
\end{table*}

\section{Algorithm Pseudocode}

\begin{algorithm*}[t]
\caption{HiSME Online Skill Meta-Evolving}
\label{alg:hisme}
\begin{algorithmic}[1]
\AlgInput training tasks $X_{\mathrm{train}}=\{x_1,\ldots,x_T\}$, frozen executor $\mathrm{Exe}_\phi$, initial library $S_0$, role rules $M_0$
\AlgOutput active skill library $S_T^{\mathrm{active}}$ and role rules $M_T$
\State Initialize credit table $C$, overlap graph $G$, and role buffers $\mathcal{D}^{\mathrm{ext}},\mathcal{D}^{\mathrm{rfg}},\mathcal{D}^{\mathrm{rfn}}$
\For{$t=1$ to $T$}
    \State $\tau_t \gets \Call{ExecuteWithRetrievedSkills}{\mathrm{Exe}_\phi,x_t,S_{t-1},K}$
    \State $U_t \gets \Call{EvaluateTrace}{x_t,\tau_t}$
    \State $z_t \gets \Call{Project}{\tau_t,U_t}$
    \State $G \gets \Call{UpdateOverlapGraph}{G,z_t,S_{t-1}}$
    \State $E_t \gets$ skills retrieved, injected, or explicitly used in $\tau_t$
    \State $Y_t \gets \Call{AssignCredit}{\tau_t,U_t,E_t}$
    \State $C \gets \Call{UpdateCreditTable}{C,Y_t}$
    \State $G_t^{\mathrm{ext}} \gets \Call{ExtractCandidates}{\tau_t,S_{t-1},M_{\mathrm{ext}}}$
    \State $S_t \gets S_{t-1}\cup G_t^{\mathrm{ext}}$
    \If{$t \bmod k_{\mathrm{micro}}=0$}
        \State $S_t \gets \Call{MicroMaintain}{S_t,C,Y_t,M_{\mathrm{rfn}}}$
    \EndIf
    \If{$t \bmod k_{\mathrm{macro}}=0$}
        \State $S_t \gets \Call{MacroMaintain}{S_t,G,C,M_{\mathrm{rfg}},M_{\mathrm{rfn}}}$
        \State $(\mathcal{D}^{\mathrm{ext}},\mathcal{D}^{\mathrm{rfg}},\mathcal{D}^{\mathrm{rfn}})\gets \Call{BuildRoleFeedback}{S_t,C}$
        \State $M_{\mathrm{ext}}\gets \Call{UpdateRoleRules}{\mathrm{ext},M_{\mathrm{ext}},\mathcal{D}^{\mathrm{ext}}}$
        \State $M_{\mathrm{rfg}}\gets \Call{UpdateRoleRules}{\mathrm{rfg},M_{\mathrm{rfg}},\mathcal{D}^{\mathrm{rfg}}}$
        \State $M_{\mathrm{rfn}}\gets \Call{UpdateRoleRules}{\mathrm{rfn},M_{\mathrm{rfn}},\mathcal{D}^{\mathrm{rfn}}}$
    \EndIf
\EndFor
\State \Return $(S_T^{\mathrm{active}},M_T)$
\end{algorithmic}
\end{algorithm*}

\begin{algorithm*}[t]
\caption{Execution with Retrieved Skills}
\label{alg:retrieval}
\begin{algorithmic}[1]
\AlgInput executor $\mathrm{Exe}_\phi$, task $x$, skill library $S$, retrieval size $K$
\AlgOutput trace $\tau$
\State $\tau\gets \emptyset$
\For{turn $j$ in $x$}
    \State $q_j\gets \Call{BuildRetrievalQuery}{x,j,\tau}$
    \State $R_j\gets \operatorname{TopK}_{s\in S_{\mathrm{active}}}\rho(q_j,s)$
    \State $p_j\gets \Call{NativePrompt}{x,j,\tau}\oplus \Call{RenderSkills}{R_j}$
    \State $(a_j,e_j)\gets \mathrm{Exe}_\phi(p_j)$
    \State $\tau\gets \tau\cup\{(j,q_j,R_j,a_j,e_j)\}$
\EndFor
\State \Return $\tau$
\end{algorithmic}
\end{algorithm*}

\begin{algorithm*}[t]
\caption{Overlap-Graph Refactoring}
\label{alg:overlap}
\begin{algorithmic}[1]
\AlgInput graph state $G$, skill library $S$, refactorer rules $M_{\mathrm{rfg}}$
\AlgOutput updated skill library $S'$
\State $V\gets$ trace segments and active or pending skill nodes
\ForAll{candidate pairs $(u,v)\in V\times V$}
    \State $w(u,v)\gets \alpha\,\mathrm{sparse}(u,v)+\beta\,\mathrm{emb}(u,v)+\gamma\,\mathrm{err}(u,v)$
    \If{$w(u,v)\ge\eta$}
        \State add weighted edge $(u,v)$ to $G$
    \EndIf
\EndFor
\State $\mathcal{C}\gets \Call{StrictCliqueSearch}{G,C_{\min},C_{\max}}$
\ForAll{$C\in \mathcal{C}$ with explainable purity}
    \State $P_C\gets$ clique evidence, selected segments, related skills, and $M_{\mathrm{rfg}}$
    \State $G_C\gets \Call{LLMRefactorer}{P_C}$
    \ForAll{$s\in G_C$}
        \If{$s$ is supported by all true instances and $\Call{Gate}{s}$ passes}
            \State add $s$ to $S$ with source role \textsc{refactorer}
        \EndIf
    \EndFor
\EndFor
\State \Return $S$
\end{algorithmic}
\end{algorithm*}

\begin{algorithm*}[t]
\caption{Credit, Bundle Maintenance, and Refinement}
\label{alg:maintenance}
\begin{algorithmic}[1]
\AlgInput trace $\tau$, metrics $U$, exposed skills $E$, skill library $S$
\AlgOutput updated skill library $S'$ and credit table $C'$
\State $Y\gets \Call{LLMCreditAssigner}{\tau,U,E}$
\ForAll{$y_{t,s}\in Y$}
    \State update exposure, usage, helpful, harmful, neutral, and uncertain counts in $C_s$
    \If{$y_{t,s}$ implies a new bundle case}
        \State patch $B(s)$ with a minimal positive, negative, or integration fragment
    \EndIf
    \If{$y_{t,s}$ or bundle failure requires refinement}
        \State $G_{\mathrm{rfn}}\gets \Call{LLMRefiner}{s,B(s),C_s,M_{\mathrm{rfn}}}$
        \ForAll{$s'\in G_{\mathrm{rfn}}$}
            \State reset $C_{s'}$, record parent$(s')=s$, and run $\Call{Gate}{s'}$
        \EndFor
        \State replace, rollback, disable, or archive $s$ according to revision evidence
    \EndIf
\EndFor
\ForAll{skills $s$ with sufficient harmful evidence and insufficient helpful evidence}
    \State disable or archive $s$
\EndFor
\State \Return $(S,C)$
\end{algorithmic}
\end{algorithm*}

\begin{algorithm*}[t]
\caption{Role-Specific Meta-Rule Update}
\label{alg:meta}
\begin{algorithmic}[1]
\AlgInput role $r$, current rules $M_r$, role feedback buffer $\mathcal{D}^r$
\AlgOutput updated role rules $M'_r$
\State $\mathcal{E}\gets \{d_s\in\mathcal{D}^r:\mathrm{mature}(s)\}$
\State $F\gets$ compact evidence from $\mathcal{E}$, including skill signatures, credit, bundle outcomes, lifecycle decisions, and parent--revision comparisons
\State $R\gets \Call{LLMMetaMaintainer}{r,M_r,F}$
\If{$R$ contains parseable analysis, summary, and rule sections}
    \State $M'_r\gets$ at most five normalized concise rules from $R$
\Else
    \State $M'_r\gets M_r$
\EndIf
\State \Return $M'_r$
\end{algorithmic}
\end{algorithm*}

\end{document}